# Towards automatic extractive text summarization of A-133 Single Audit reports with machine learning

Vivian T. Chou
*Harvard Medical School*
Boston, MA, USA
vtchou@gmail.com

LeAnna Kent
*Elder Research, Inc.*
Arlington, VA, USA
leanna.kent@elderresearch.com

Joel A. Góngora
*Elder Research, Inc.*
Arlington, VA, USA
joel.gongora@elderresearch.com

Sam Ballerini
*Elder Research, Inc.*
Arlington, VA, USA
sam.ballerini@elderresearch.com

Carl D. Hoover
*Clarkson University*
Potsdam, NY, USA
choover@clarkson.edu

*Abstract*—The rapid growth of text data has motivated the development of machine-learning based automatic text summarization strategies that concisely capture the essential ideas in a larger text. This study aimed to devise an extractive summarization method for A-133 Single Audits, which assess if recipients of federal grants are compliant with program requirements for use of federal funding. Currently, these voluminous audits must be manually analyzed by officials for oversight, risk management, and prioritization purposes. Automated summarization has the potential to streamline these processes. Analysis focused on the "Findings" section of ~20,000 Single Audits spanning 2016-2018. Following text preprocessing and GloVe embedding, sentence-level *k*-means clustering was performed to partition sentences by topic and to establish the importance of each sentence. For each audit, key summary sentences were extracted by proximity to cluster centroids. Summaries were judged by non-expert human evaluation and compared to human-generated summaries using the ROUGE metric. Though the goal was to fully automate summarization of A-133 audits, human input was required at various stages due to large variability in audit writing style, content, and context. Examples of human inputs include the number of clusters, the choice to keep or discard certain clusters based on their content relevance, and the definition of a top sentence. Overall, this approach made progress towards automated extractive summaries of A-133 audits, with future work to focus on full automation and improving summary consistency. This work highlights the inherent difficulty and subjective nature of automated summarization in a real-world application.

## Introduction

Text summarization is the process of shortening long pieces of text into a coherent summary containing only the main points of the document. Summarization is a subfield of natural language processing (NLP) that has garnered increasing attention over recent years due to the rapid growth of text-based data. While automated summarization methods have been utilized since the mid-20th century to generate abstracts of domain-specific technical papers [1]–[3], human-written summaries have remained largely sufficient for many applications for many decades. Early NLP researchers foresaw that the continued growth in knowledge would necessitate efficient and consistent summarization on a scale that could not be achieved manually [4]. Indeed, the "information explosion" of the 21st century has stimulated broad use of automated summarization not just on technical documents but on a far greater variety of texts, both in terms of subject and format.

A summary can be defined as "…a text…that conveys important information in the original text(s), and that is no longer than half of the original text(s) and usually significantly less than that" [5]. The abstract of this paper is a simple example of a summary. Summarization can be achieved through two major approaches, extractive and abstractive [6]. Extractive summarization selects important sentences/phrases from the original text and combines them, verbatim, into a summary, while abstractive summarization generates sentences *de novo*. Abstraction more closely resembles human cognition but is very challenging to implement and does not necessarily produce clearly superior summaries. Despite the relative simplicity of extractive summarization, it can produce highly effective summaries and is widely utilized. We will thus focus on extractive summarization, and any references hereafter to "summarization" will refer to extractive methods (although many ideas are also applicable to abstractive summarization).

One readily applied method of summarization is latent Dirichlet allocation (LDA) [7], which represents a corpus as a collection of topics or words. While the generative methodologies and assumptions of LDA make it very flexible, its ability to produce informative summaries may be limited by its definition of a "topic," therefore motivating the use of alternative strategies in certain contexts. Here, we describe one alternative that utilizes word embedding and clustering. The bulk of this work, like many NLP workflows, lies in the preprocessing methodology; we present a method in which sentences/phrases of the corpus are embedded into single multi-dimensional mean vector representations and grouped by cosine similarity. Finally, for each document in the corpus, the sentences/phrases nearest to the centroid of each group are defined as those which are most important to the document.

## I. Extractive Summarization Methodology

For any endeavor, the specific summarization strategy will vary depending on the source text, the desired output, and available resources, among other variables. Mechanistic details aside, at the fundamental level, nearly all extractive summarization approaches are built upon a fairly universal set of principles: 1) initial conversion of text to an intermediate *numerical representation*, followed by 2) the *ranking and selection* of the most important units (sentences, paragraphs, etc.). In practice, the workflow may not break down so cleanly or discretely. For instance, a single technique may accomplish both numerical representation and ranking. On the other hand, any given "step" might encompass multiple techniques, e.g. a combination of algorithms may be used to rank and select sentences. Steps may be further subdivided, e.g. separate ranking and selection. Small variations notwithstanding, virtually all summarization is predicated on these major tasks.

The typical first step, *numerical representation*, is necessary to convert text to a computer-readable format and calculate quantitative features for downstream algorithms. Pioneering summarization works utilized statistical calculations such as word/phrase frequency [1], sentence position [3], the presence of key phrases [2], and/or term frequency-inverse document frequency (TF-IDF) [8]. Nowadays, these older methods remain in use, though more sophisticated methods may be preferred. One popular approach is to represent words as vectors, i.e. "word embeddings." While the concept of word embeddings dates back to the 1990s and 2000s [9][10], the most significant breakthroughs arguably occurred in 2013-2014, with the introduction of Word2Vec [11][12] and GloVe [13]; these embeddings have since become mainstays of not only summarization but of NLP. Very recently, novel embedding methods such as ELMo [14], BERT [15], and XLNet [16] have provided innovative alternatives.

Following *numerical representation*, sentences must be *ranked* on the basis of importance, and the "top" sentences are *selected*. If the data are labeled, supervised or semi-supervised machine learning can be used to train a classifier that predicts the importance of a sentence [17]. While supervised and semi-supervised learning can achieve very desirable results, many datasets are unlabeled, largely precluding these methods. It is possible to hand-label the data; however, the amount of labeled training data needed generally makes this approach very laborious even for semi-supervised methods that do not require as much labeled data. Thus, extractive summarization often relies instead on unsupervised learning. Such approaches include graph-based algorithms, such as the popular TextRank [18], which has its roots in Google's PageRank algorithm [19], as well as unsupervised deep learning [20][21]. Clustering methods, such as *k*-means [22], are another broad category of unsupervised methods used in summarization [23]–[28].

## II. Summary Evaluation Approaches

Beyond the task of summarization per se, the problem of how to evaluate the quality of the eventual output warrants careful consideration. Challenges arise not only from the concrete implementation of the evaluation method but also from the broader question of what "quality" entails. Such questions include: How does one resolve the trade-off between sufficient brevity and content? How important are factors such as readability and coherence, alongside essential elements such as grammar? Naturally, the answers to these questions are not only subjective but also highly circumstantial, as they will be influenced by the intended purpose of the summaries, among other variables. The actual utility of the summary to the end user might be the most decisive factor, but this information may only become available on a longer timescale.

Evaluation is also complicated by the general requirement for significant human input. For instance, the DUC and TAC summarization conferences have employed human judges to grade summaries; however, such labor-intensive approaches are not broadly practical. Alternatively, automated evaluation methods are implemented by packages such as Recall-Oriented Understudy for Gisting Evaluation, or ROUGE [29] and its precursor, BLEU [30], which calculate precision and recall scores that reflect the quality of a summary. However, ROUGE and similar methods may still require considerable human effort, as the system-generated summary must be compared to a human-written reference summary to calculate the required metrics. Thus, the utility of ROUGE is somewhat conditional, i.e. it may only be suitable for a corpus where each document is accompanied by a pre-existing abstract (as with most academic journal articles), abstract-like text (introductory blurbs for Wikipedia articles), or additional metadata. ROUGE is also contingent on the quality of the reference summary, which is itself subject to all the questions and criteria outlined above about what constitutes a summary of good quality.

In recognition of these limitations, alternate evaluation techniques have been developed, e.g. using latent semantic analysis (LSA) to measure the similarity of system-generated summaries to the original source text [31]. Ultimately, the problem of evaluation remains a many-headed hydra, and as with any other aspect of automatic summarization, the "right" technique is likely to be highly context-specific.

## III. Summarization of A-133 Single Audits, a Rich Resource for Federal Grant Agencies

Our current work focuses on summarization of a corpus of A-133 Single Audits from FY 2016-2018 to capture the qualities and activities of grant recipients detailed in the audits. There are significant financial and economic motivations for better understanding grant recipients. Grants are one of the largest categories of spending by the United States government, which is projected to award over $750 billion in grants in FY 2019. To ensure proper use of funds, Single Audits (full database at harvester.census.gov/facweb/) are conducted on all organizations that expend at least $750,000 of federal grants in a year to assess program compliance by analysis of a recipient's financial records, financial statements, federal award transactions and expenditures, internal control systems, and the federal assistance received during the audit period. Thus, the Single Audit corpus represents a valuable resource for decision-making and risk management by grant agencies. For instance, grants managers utilize the information within Single Audits to decide whether or not to award a grant based on an applicant's past performance; or, if a grant has already been awarded, when and how to perform oversight.

Manually extracting information from Single Audits is nontrivial, as the length of the audits makes it cumbersome for a human reader to locate the desired information. It would be beneficial to identify and frame the most relevant information in a compressed, easily digestible format. The volume of the data presents an additional barrier. While the corpus used in this study spans only FY 2016-2018, it already comprises over 40,000 documents (~2.5 million pages) and will inevitably grow over time. Thus, automation is not only necessary to accomplish corpus-wide analyses on a reasonable timescale but also to ensure consistent and unbiased outcomes.

*A. Related work on A-133 Single Audits*

The current work builds upon an earlier project at Elder Research, Inc., which devised an automated method to quantify the severity of an audit and an associated numerical risk score. To this end, Single Audits from 2016-2018 were downloaded to form a corpus. While most audits are 45-60 pages long, and some are hundreds of pages, the pertinent information is contained in the section titled "Schedule of Findings and Questioned Costs," which typically spans a few pages. A method was devised to extract only the relevant "Findings" pages, which comprise ~1.5% of the pages in the corpus. This enables the analysis to focus solely on the most essential data. Sentiment analysis of the extracted Findings was performed to generate a risk score for each audit report. Altogether, this work makes it possible to quickly assess grantees through a straightforward, intuitive metric and facilitates prioritization of the grantees of most interest, e.g. those deemed most "risky."

*B. Motivation for and overview of current approach*

The previous work by Elder Research offers a powerful method to answer the question of *who* is high risk, and raises the natural question of *why* these entities are high risk. Here we attempt to address this latter question of *why* by extracting the text that provides explanations. The choice to answer *why* through automatic summarization, as opposed to simply using manual summaries, is driven by the volume of data. Our corpus of selected 2016-2018 audits already comprises 40,000 documents, and the problem only exacerbates when one considers the rest of the audits in existence, to say nothing of the new documents that will be added year after year. Even if one were only interested in a much smaller subset of documents for which manual summarization is reasonable, it is unlikely that a human (or group of humans) would maintain consistent quality and objectivity over so many documents.

For our current approach, the desired output for each document is a summary consisting of the sentence(s) that best illustrate the broader context. Below are examples of sentences that would be desirable to extract because they help elucidate the reasoning behind a negative assessment of a grantee:

"Grantee was unable to provide proper documentation."

"A student at the school was missed and never corrected."

"The cause stemmed from turnover within staff and ultimately from a lapse in procedures."

Once the relevant documents were accessed, the data were prepared for modeling. Preparation entailed both standard and corpus-specific text preprocessing steps, followed by word and sentence embeddings using pre-trained GloVe vectors. Sentence vectors from the entire corpus were pooled and used to train a *k*-means clustering model, which was iteratively optimized. Summaries for each document were generated by extracting the sentences deemed most important and subsequently evaluated using both human and system metrics.

Our initial task was to extract the Findings sections and exclude all other pages in the audits to reduce noise. This a excluded a substantial number of entire reports that did not contain Findings, which typically pertained to high-performing grantees that did not elicit critique The corpus was thus reduced from ~40,000 full-length reports to 19,234 abbreviated documents containing only Findings. Any reference hereafter to "document" will refer to the abbreviated documents. While this filtering approach helps simplify the problem, it may induce a selection bias in subsequent analysis that can only be evaluated with out-of-sample performance testing.

IV. DATA EXPLORATION

Each document was initially represented as a single string, which was tokenized, i.e. divided or split, on a sentence- or word-level as needed. To ensure that all canonically equivalent strings have the same binary representation, NFKD Unicode normalization was performed on all text prior to preprocessing.

To gain an intuition for the corpus, simple univariate analysis of document lengths was performed. First, each string for each document was divided into its constituent sentences using the Natural Language Toolkit (NLTK) sentence tokenizer; note that tokenization at this stage was only for the purposes of measuring document length, and that the un-tokenized string was used for eventual preprocessing. Sentence counts were analyzed, yielding some notable observations:

- The mean document length was 30 sentences.
- The corpus is skewed towards shorter documents: the mode was 3 (sentences), followed by 4 and 1.
- 96% of documents were 100 sentences or shorter.
- 0.6% of documents were "extra-long, >500 sentences.
- The longest document was 3940 sentences.

We randomly selected and read a fixed number of documents from each length bracket (short, medium, long, extra-long) to gain a sense for organization and content as well as orthography and stylistic conventions. Some observations:

- Most documents contained brief filler sentences, such page headers/footers, e.g. "State of New Mexico," "Denver Housing Center"; section headings, e.g. "Condition," "Cause"; and page numbers.
- Many documents contained expository statements (e.g. historical background) not relevant to risk. Such statements predominated in longer documents and "dilute" the relevant information; information was more concentrated in short/medium documents.

- The shortest documents corresponded to grantees that had satisfactory audits, and might only contain 1-2 sentences, e.g. "No findings or questioned costs."
- There was a lack of universal formatting, which proved challenging during analysis. Section headings could appear as "Cause:" or "CAUSE" and inline or separately; lists might appear as "1." or "(1)".

## V. TEXT PREPROCESSING

We performed a combination of standard and corpus-specific preprocessing (Table I) to ensure predictable and consistent results. Standard preprocessing reduces noise associated with virtually any text data, such as punctuation and frequently-occurring stopwords. Due to the specialized nature of Single Audits, our corpus also contained idiosyncrasies in diction, syntax, format, and structure that warranted additional preprocessing and/or modification of the standard steps.

Manual preprocessing occurred in two major stages (Fig. 1). Stage A removed clearly irrelevant sentences but preserved orthography to generate human-readable intermediate text. Stage B produced fully-processed sentences for embedding. Thus, while ranking/selection was performed on the fully-processed sentences, the corresponding intermediate sentences were used to produce a readable summary output.

### A. Initial preprocessing to generate readable intermediates

At the outset, each document was represented by a single string. Page headers and section headings were removed based on length: we looked for all substrings that occurred between two newline breaks and removed substrings less than 60 characters. To avoid removal of dangling lines, we only removed short substrings that started with an uppercase letter or a non-alphabetical character. Each document was then divided using the NLTK sentence tokenizer. To further isolate section headings, each sentence that contained a colon or dash was further split at that character. Additional elements were removed using regular expressions ("regex," Table II).

TABLE I. STANDARD VS. CORPUS-SPECIFIC PREPROCESSING

| Standard preprocessing | Corpus-specific |
|---|---|
| No stemming/lemmatization or n-gram modeling performed | Remove page headers and short sentences |
| Convert text to lower-case | Remove Roman numerals |
| Remove punctuation, symbols, extra whitespace | Remove parenthetical phrases |
| Remove stopwords (NTLK list) | Keep negation stopwords (*no, not*) |

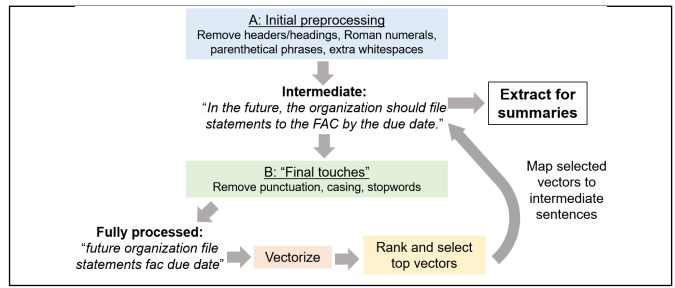

Fig. 1. Breakdown of preprocessing stages and subsequent steps.

TABLE II. REGULAR EXPRESSIONS FOR TEXT CLEANING

| Stage | Removal Criteria | Regular Expression |
|---|---|---|
| A | Roman numerals | ^(?=[MDCLXVI])M*(C[MD]\|D?C{0,3})(X[CL]\|L?X{0,3})(I[XV]\|V?I{0,3})$ |
| | Parenthetical phrases | [(].+[)] |
| | Extra whitespace | \s+ |
| B | Non-alphanumeric | [^a-zA-Z0-9]+ |

Each document was stored as list of strings identifiable by index. The intermediate text still contained typical English orthography, e.g. capitalization, punctuation, and stopwords. Though these sentences were not suitable for embedding, they remained easily readable and were retained for later use.

### B. "Final touches" and sanity check

We next lower-cased the entire text and removed punctuation (using regex, Table II). The final step, stopword removal, was performed in tandem with GloVe embedding (see next section). Stopword removal can be accomplished using the predefined NLTK Stopword list. For this corpus, one key modification was to retain the words *no* and *not*. While these words may not add value to certain corpora, they were important for this corpus because audits often focus on the lack or failure to fulfill a certain requirement; these words were also important to determining if no problems were detected.

After preprocessing, 4,233 documents (22% of the corpus) had zero length, i.e. became empty lists. To ensure that preprocessing was not removing key information, the original vs. preprocessed text lengths were compared. The rationale for was that if the preprocessing were functioning properly, empty lists should originate from documents that were short and/or had little substantial content. As expected, the documents that produced empty lists were 13 sentences or shorter originally, and were ~3 sentences long on average. Even the longest of these documents contained only filler text, such as headers. These results indicated that preprocessing was sensible and that empty lists were valid reflections of the original data, rather than artifacts of over-exuberant preprocessing. Though necessary in this case, all such filtering of information is non-ideal for an automated process and may induce selection bias.

The remaining 15,001 documents (78% of the corpus) stayed non-empty after preprocessing. Interestingly, 4% of the corpus showed increased sentence counts, likely from splitting of longer sentences. This was unlikely to be problematic, since information was merely partitioned and not lost. Regardless, this length increase was uncommon, and 74% of the corpus showed reduced length following preprocessing, with the cleaned text about ~70% the sentence count of the original text.

## VI. WORD/SENTENCE EMBEDDING WITH GLOVE

Upon completion of preprocessing (stages A and B), the fully-cleaned corpus was ready for embedding. The GloVe method was chosen based on its recognized effectiveness across many NLP tasks as well as prior success in using GloVe with this corpus. The ease and speed of implementation of pre-trained vectors was also a decisive factor due to time constraints. A natural extension would be to use methods, such as BERT or custom GloVe, that require modifications.

The "glove.6B.zip" file containing vectors trained on the English-language Wikipedia 2014 + Gigaword 5 corpus was downloaded (from nlp.stanford.edu/projects/glove/). We obtained individual word tokens for word-level embedding by splitting sentences at each whitespace. For each non-stopword word, the appropriate word vector was retrieved. A single vector for each sentence was then obtained by taking the arithmetic mean across each element of the constituent word vectors. A vector of all zeros was automatically assigned to the very rare sentences comprised solely of 1) stopwords or 2) an empty string; this occurred in only 0.02% of the corpus (96 sentences out of 452,071). This entire embedding process, and subsequent $k$-means clustering (see next section) was performed using both 50d and 100d vectors. Since 100d vectors conferred no obvious advantage, the less memory-intensive 50d vectors were used for the remainder of the work.

## VII. Sentence ranking/selection by $k$-means

Both ranking and selection of the vectorized sentences were accomplished by $k$-means clustering, which was chosen for its intuitive interpretation in this context. The algorithm partitions $n$ observations, i.e. sentences ($n$ = 452,071) into $k$ clusters, i.e. topics, by assigning each observation to the nearest centroid (Fig. 2). The centroids and clusters are iteratively updated to minimize intra-cluster variation, i.e. the within-cluster sum of squares. The importance of each sentence correlates to its proximity to its cluster centroid, allowing identification of the top sentences per cluster/topic (Fig. 2A) and overall (Fig. 2B).

### A. Implementation details and hyperparameter tuning

To train the $k$-means model on the entire corpus at once, vectorized sentences were pooled and stored in a flattened list. Each vector was mapped back to its text representation based on its unique index in the list. Clustering was performed using both the scikit-learn and NLTK implementations of $k$-means.

#### 1) Selection of cosine similarity as the distance metric

A key $k$-means hyperparameter is the distance metric used in assigning centroids and clusters. While Euclidean distance (1) is a common default, cosine similarity (2) is preferred for text and therefore was our metric of choice for all modeling and calculations. The scikit-learn $k$-means implementation does not allow ready use of a metric besides Euclidean distance. To circumvent this, we trained our scikit-learn $k$-means model on vectors normalized to unit length, based on the principle that squared Euclidean distance and cosine similarity are proportional for unit vectors (3). Because the NLTK $k$-means implementation allows specification of the distance metric, normalization was only necessary for the scikit-learn model.

Denote squared Euclidean distance:
$$\|x-y\|_2^2 \quad (1)$$

And the cosine metric:
$$\cos\angle(x,y) \quad (2)$$

By expansion of squared Euclidean distance,
$$\|x-y\|_2^2 = (x-y)^T(x-y) = x^Tx - 2x^Ty + y^Ty = 2 - 2x^Ty + y^Ty$$
$$= 2 - 2x^Ty = 2 - 2\cos\angle(x,y) \quad (3)$$

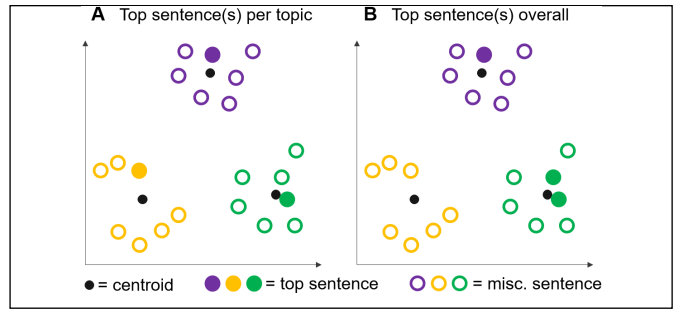

Fig. 2. Simplified visualization of k-means clustering to rank and select sentences. Color-coded clusters correlate to distinct topics. Top sentences were selected by (A) importance within individual topic or (B).

#### 2) Selection of cluster number, $k$

The other major hyperparameter was $k$, representing the number of clusters/topics. We first utilized the "elbow method," whereby $k$-means is run for a range of $k$ and the sum of squared errors for each model is plotted as a function of $k$. To reduce spatial/temporal costs, the scikit-learn Mini-Batch $k$-means model was run on $k$ = [2, 30). The resulting plot indicated an optimal $k$ around 12-14. These values were tested empirically by running the normal scikit-learn $k$-means model on cluster sizes of 12, 13, and 14, with 12 performing the best.

#### 3) Ranking and selection of sentences

The $k$-means algorithm outputs $k$ centroids, which represent the means of all observations in a given cluster. The distance of each observation to its respective centroid can then be calculated. We thus generated a matrix of observation-centroid distances. For a given document, top sentences were selected by cluster (Fig. 2A) or by overall proximity (Fig. 2B). These processes can be described algorithmically:

- *Top sentences by cluster.* Consider all sentences in cluster 1. If no sentences belong in this cluster, go on to the next cluster; else, rank sentences by proximity to cluster 1 centroid and save the index of the top sentence(s). Repeat for clusters 2-$k$. The result is a list of sentence indices, of length $k$ or less.

- *Top sentences overall.* Begin with the matrix storing the proximity of each sentence in the document to its cluster centroid. Sort sentences by distance. Select top $j$ sentences. The result is a list of sentence indices, of length $j$ or the document length, whichever is smaller. $j$ was roughly scaled to the original document length:

    - < 15 sentences, $j$ = length/2 (rounding up).
    - 15-80 sentences, $j$ = 7.
    - 80-250 sentences, $j$ = 12.
    - > 250 sentences, $j$ = length/20 (rounding up).

These values were chosen empirically to attempt to balance sufficient information with brevity and to obtain summaries roughly the same length as those obtained by the per cluster method. The output for each document was a list of indices, and the output for the corpus was a list of lists. Summaries were generated by extracting the intermediate sentence (see preprocessing Stage A) corresponding to each of the indices.

## B. Preliminary results and filtering of clusters by relevance

As desired, a text summary was successfully generated for each document following these steps. Preliminary evaluation (via reading randomly sampled summaries) revealed that despite preprocessing, summaries still contained filler text, such as page headers or section headings. The persistence of such text despite preprocessing was likely due to inconsistent formatting across documents, preventing detection of filler text by established criteria. Summaries also contained extraneous text that were valid sentences and therefore would not have been filtered out by preprocessing, but that nonetheless added little value to the summaries. These sentences included (but were not limited to) logistical information such as lists of dates and the amounts of individual financial transactions; while this information may have value in certain contexts, they do not greatly improve understanding of grantee risk. It would be preferable to omit such sentences from the eventual summaries.

To reduce the amount of extraneous information in the summaries, the 12 clusters were selectively discarded based on content. Two criteria were used to filter the clusters:

- *Cluster topic*, determined by examining the 10-20 sentences closest to the centroid.
- *Uniformity of content*, determined by examining a random sample of 10-20 sentences from each cluster.

Ultimately, 5 out of 12 clusters (produced with both scikit-learn and NLTK models) were discarded, leaving 7 "useful" clusters of the original. For certain clusters, this decision was straightforward: for instance, both scikit-learn and NLTK models produced one cluster (out of 12) that contained exclusively page headers and section headings, as well as a cluster that contained almost exclusively dates. Other discarded clusters were characterized by excessive financial detail, e.g. dollar amounts of transactions; overly long sentences; and sentences that are part of the standard verbiage of audits but that do not add information specific to individual grantees.

The choice to keep or discard certain clusters was, in some cases, not straightforward. While some clusters were almost entirely composed of irrelevant text, this was not universal: the content of most of clusters was far from uniform. Clusters which contained predominantly irrelevant information would also contain a non-trivial quantity of relevant; similarly, clusters that were "useful" as a whole would still contain many sentences of little value. Arguably, discarding any but the most unambiguous clusters risks significant loss of information. This is further complicated by the question of what is "relevant" or "irrelevant." For instance, we decided that sentences stating exact dollar amounts of transactions were excessively detailed. However, it could be argued that financial details do provide clues to a grantee's risk and should be retained.

In short, cluster filtering is a challenging step that is rife with ambiguities and trade-offs, and the current manual approach has much potential for improvement, such as a more thorough analysis of intra- and inter-cluster distances and of cluster density via silhouette scores. Given the major effect of cluster filtering on summary quality (see next section), greater investment on examining the different clusters and on methods to prioritize information would likely have significant benefit.

|  | | *k*-means implementation | |
|---|---|---|---|
|  |  | sci-kit learn | NLTK |
| Selection of top sentences | Per Cluster | A scikit-learn per cluster | B NLTK per cluster |
|  | Overall | C scikit-learn overall | D NLTK overall |

Fig. 3. Distinct summaries produced by different combinations of *k*-means implementations and sentence selection methods.

Throughout the above steps, the text output was evaluated by ad hoc read-throughs by a graduate-level data scientist/ analyst, not a domain expert in audits. Even from preliminary evaluation, it was apparent that removal of irrelevant clusters significantly changed the summary content. Despite concerns about information loss, the summaries as a whole appeared to have a greater proportion of relevant content. Thus, this change was implemented despite potential trade-offs. To mitigate possible loss of information from the "per cluster" method due to the reduction of available clusters from 12 to 7, two sentences (rather than one) were chosen per cluster; the number of sentences selected "overall" still followed the same criteria.

Four sets of summaries were ultimately generated (Fig. 3). These summaries were evaluated both in isolation for their overall quality, as well as relative to each other to determine if any method conferred obvious advantages. To this end, 20 "medium" length documents (15-100 sentences prior to summarization) were chosen randomly. These 20 documents were then evaluated both manually and using the ROUGE metric. In general, the four summary sets appeared comparable, with the "per cluster" methods performing slightly better (Fig. 3A,B); there was no conspicuous advantage to either the scikit-learn (Fig. 3A,C) or the NLTK implementations (Fig. 3B,D).

## VIII. HUMAN EVALUATION

Manual evaluation was performed with these criteria:

- *Brevity*. Were the summaries of a reasonable length for high-throughput reading? While the selection criteria imposed a length limit, summaries might still be longer than desired. For instance:
  - A 14 sentence summary is not impressive if the original text was 16 sentences.
  - Occasional very long sentences also inflated summary lengths in an unpredictable way.
- *Information*. Did the summary contain enough information to understand grantee risk? Did the summary successfully omit unnecessary information?
- *Explicit wording*. How transparent and accessible was the information?

Besides these benchmark questions, no formal rubric or scoring system was used; thus, human evaluation relied largely on an intuitive sense of "good" and "poor."

We found that the sampled summaries were of varying quality. Best-case summaries fulfilled all three above criteria. Worst-case summaries were composed entirely of irrelevant information; such summaries would be functionally useless no matter how well they fulfilled any other criteria. Most commonly, our summaries were functional even if not optimal. Three broad scenarios (not mutually exclusive) were apparent:

- *Multiple findings*. A single summary contained multiple distinct findings, resulting in a logically confusing text (more details in final section).
- *Dilution of information*. The desired information was present but was "hidden" among sentences that were not desired, thus requiring the reader to actively filter out the unnecessary information.
- *Oblique phrasing*. Ideally, the desired information would be stated explicitly, e.g. "The grantee was non-compliant because [some action] was not performed." In practice, summaries contained statements such as, "It is recommended that grantees perform [some action] to be compliant." It is possible to infer the former statement from the information in the latter, but this would require additional effort on the part of the reader, which may or may not be tenable.

Summaries pertaining to the latter two scenarios might be acceptable, since they still contain the desired information, with the caveat that they would likely yield a suboptimal user experience. The first scenario of multiple findings (see final section for more detail) is perhaps the most problematic because of the potential for missing or misleading information. Due to the significant challenges of separating documents by findings, the problem of multiple findings was not resolved in this work (but would be a valuable future endeavor).

## IX. ROUGE EVALUATION

The 20 summaries evaluated manually were also evaluated using ROUGE. For each summary, a human reference was written. Overlaps between the system generated summaries and corresponding "ideal" human reference were then calculated. To reduce noise in the calculations, the above-mentioned preprocessing steps of stopword filtering, punctuation removal, and lower-casing were performed on both the human and system summaries; stemming was also done using the NLTK PorterStemmer. ROUGE produced two key metrics:

- *Recall*. How much of the human reference summary was captured by the system summary?
- *Precision*. How much of the system summary was actually needed or relevant?

Despite the popularity and general effectiveness of the ROUGE method, for this particular application, human evaluation was considered more informative than ROUGE. Some shortcomings of ROUGE could be attributed to the general pitfalls of the method, although a few additional issues were noted. On the whole, the recall and precision scores were poor (~50%) compared to what is typically desired (70-90%), making it difficult to confidently assess summary quality. Another shortcoming was that only one set of reference summaries was available, which constricted the definition of "ideal." In reality, there is rarely a single ground truth, and a text can have many valid summaries; thus, the use of multiple reference summaries, written by different people, would likely improve ROUGE evaluation. Finally, since ROUGE was only performed on 0.1% of the corpus and moreover limited to documents of a certain length (15-100 sentences), it might be ill-advised to extrapolate these results to any significant extent.

We have provided potential future directions based on our current assessment of our method. It will also be invaluable to solicit expert user feedback to ensure that our method achieves not only technical soundness but also its desired functionality.

## X. SHORT-TERM MODIFICATIONS

These comparatively minor modifications do not radically change the overarching logic of the approach.

### A. Word embeddings

We have noted several alternatives to GloVe (see Methods section). While it is difficult to precisely predict the results of any method and how much they would improve the output (if at all), we speculate that customized embeddings would better capture the specialized language of audits. The GloVe vectors currently in use were trained on Wikipedia and Gigaword, a newswire corpus; audit documents differ greatly from both these texts. Thus, it is likely that the words meanings and contexts captured by the pre-trained GloVe vectors do not fully reflect the usage of the words in audits. Custom vectors, e.g. LDA vector representations, may alleviate this problem.

### B. Tailor cluster number and filtering to document length

We built and uniformly applied a single *k*-means model to all documents in the corpus. However, this is almost certainly not optimal. A *k* of 12, corresponding to 12 topics, might be a decent approximation of the number of topics for an average length document of 30 sentences; it is unlikely to be the appropriate number for a document of nearly 4,000 sentences. In some cases, more clusters are likely needed to capture the different topics. Furthermore, the current method uses a narrow range of output lengths for a broad range of raw texts. While a 6-12 sentence summary might be desired for a document of 30 sentences, it is barely a summary if the original was 8 sentences and is too brief for a document that is hundreds or thousands of sentences. In the future, we will train multiple models and also explore other clustering algorithms, including density-based and hierarchical approaches.

## XI. LONG-TERM: SUMMARIZE BY DISTINCT FINDINGS

As noted above, one problematic scenario was when multiple distinct findings were collapsed into a single summary. In such a situation, the summary was, to the human reader, seemingly illogical and disjointed. This occurs because summaries are currently generated per document, and a single document may have multiple distinct findings. Intellectually, these findings are effectively separate despite originating from the same grant recipient, as they concern independent, largely non-overlapping causes/effects. Some possible consequences of condensing multiple findings into one summary include:

- *Misleading information.* For example, the cause of one finding may be attributed to the effect of a separate finding. Oftentimes, there is not enough context in the summary to detect when this happens.

- *Lack of context.* Summaries become "spread too thin" over multiple findings. For instance, some findings require multiple sentences to be fully explained. A shortcoming of the current method is that only one of several relevant sentences may be selected.

Compared to other imperfect summary outcomes, the issue is particularly noteworthy because of the potential for misinformation. A summary that reads awkwardly could still be considered "successful" if its information is relevant or accurate, and a summary with no useful information is usually clearly identifiable as being non-functional. However, the misinformation that can arise from mixed findings is not always easy to detect. To address these issues, a potential future direction could be to extract distinct findings from each audit document, and to generate a single summary for a single class of findings. This approach would have the added advantage of addressing some of the concerns related to length outlined above. Realistically, extracting findings is a complex and difficult task that would require investment of significant time and effort, but if successful, would likely have a noticeable positive effect on summarization.

## XII. Conclusion

Overall, this approach made progress towards automated extractive summaries of A-133 audits using custom text preprocessing, GloVe embedding, *k*-means clustering, and several selection heuristics. This strategy successfully extracted summaries in a technically sound algorithmic manner from a large volume of federal grant audits. Future aims include greater automation of the more exploratory steps that involved human-in-the-loop criteria. Other aims include improving the consistency of the results and further validation, both technically and in a practical setting. This work highlights the inherent difficulty and subjectivity of machine-learning based automated summarization in a real-world application and demonstrates the value of reducing non-automated steps, reducing validation subjectivity, and assessing true out-of-sample results with expert input in order to improve output.

## Acknowledgment

The authors would like to thank Robert Han and Ryan McGibony for guidance and feedback. V.T.C. would also like to thank her Ph.D. advisor David Van Vactor at Harvard Medical School for supporting her pursuit of this project, separate from her dissertation. This work was funded by Elder Research as an internal Research and Development project.

## References

[1] H. P. Luhn, "The Automatic Creation of Literature Abstracts," *IBM J. Res. Dev.*, vol. 2, pp. 159–165, 1958.
[2] H. P. Edmundson, "New Methods in Automatic Extracting," *J. ACM*, vol. 16, pp. 264–285, 1969.
[3] P. B. Baxendale, "Machine-Made Index for Technical Literature—An Experiment," *IBM J. Res. Dev.*, vol. 2, pp. 354–361, 1958.
[4] J.-M. Torres-Moreno, "Automatic Text Summarization Front Matter," in *Automatic Text Summarization*, 2014, pp. i–xxiii.
[5] D. R. Radev, E. Hovy, and K. McKeown, "Introduction to the Special Issue on Summarization," *Comput. Linguist.*, vol. 28, pp. 399–408, 2002.
[6] U. Hahn and I. Mani, "The challenge of automatic summarization," *Computer (Long. Beach. Calif).*, November, pp. 29–36, 2000.
[7] D. M. Blei, A. Y. Ng, and M. I. Jordan, "Latent Dirichlet allocation," *J. Mach. Learn. Res.*, vol. 3, pp. 993–1022, 2003.
[8] K. S. Jones, "A statistical interpretation of term specificity and its application in retrieval," *J. Doc.*, vol. 28, pp. 11–21, 1972.
[9] R. Collobert and J. Weston, "General Deep Architecture for NLP ICML 2009.pdf," 2008.
[10] Y. Bengio, R. Ducharme, P. Vincent, and C. Jauvin, "A neural probabilistic language model," *J. Mach. Learn. Res.*, pp. 1137–1155, Aug. 2003.
[11] T. Mikolov, K. Chen, G. Corrado, and J. Dean, "Distributed Representations of Words and Phrases and their Compositionality," *Neural Inf. Process. Syst.*, vol. 1, pp. 1–9, 2013.
[12] T. Mikolov, K. Chen, G. Corrado, and J. Dean, "Efficient Estimation of Word Representations in Vector Space," pp. 1–12, 2013.
[13] J. Pennington, R. Socher, and C. Manning, "Glove: Global Vectors for Word Representation," in *Proc. 2014 Conf. Empir. Methods Nat. Lang. Process. (EMNLP)*, 2014, vol. 19, pp. 1532–1543.
[14] M. Peters *et al.*, "Deep Contextualized Word Representations," in *Proc. 2018 Conf. North American Chapter Assoc. Comput. Linguist: Human Language Technologies, Volume 1 (Long Papers)*, 2018, pp. 2227–2237.
[15] J. Devlin, M.-W. Chang, K. Lee, and K. Toutanova, "BERT: Pre-training of Deep Bidirectional Transformers for Language Understanding," *arXiv Prepr.*, Oct. 2018.
[16] Z. Yang, Z. Dai, Y. Yang, J. Carbonell, R. Salakhutdinov, and Q. V. Le, "XLNet: Generalized Autoregressive Pretraining for Language Understanding," *arXiv Prepr.*, pp. 1–18, 2019.
[17] K.-F. Wong, M. Wu, and W. Li, "Extractive summarization using supervised and semi-supervised learning," *Proc. 22nd Int. Conf. Comput. Linguist.*, vol. 1, pp. 985–992, 2008.
[18] R. Mihalcea and P. Tarau, "TextRank: Bringing Order into Texts," *Proc. 2004 Conf. Empir. Methods Nat. Lang. Process.*, pp. 1–8, 2004.
[19] L. Page and S. Brin, "The Anatomy of a Large-Scale Hypertextual Web Search Engine," vol. 30, 1998.
[20] S. Verma and V. Nidhi, "Extractive Summarization using Deep Learning," *arXiv Prepr.*, Aug. 2017.
[21] M. Yousefi-Azar and L. Hamey, "Text summarization using unsupervised deep learning," *Expert Syst. Appl.*, vol. 68, pp. 93–105, Feb. 2017.
[22] J. MacQueen, "Some methods for classification and analysis of multivariate observations," *Proc. 5th Berkeley Symp. Math. Stat. Probab.*, vol. 1, pp. 281–297, 1967.
[23] D. Miller, "Leveraging BERT for Extractive Text Summarization on Lectures," *arXiv Prepr.*, 2019.
[24] N. Renu and Kunal, "Review on opinion data summarization using k-means clustering and latent semantic analysis," *Int. J. Res. Sci. Technol.*, pp. 12–23, 2017.
[25] S. Twinandilla, S. Adhy, B. Surarso, and R. Kusumaningrum, "Multi-Document Summarization Using K-Means and Latent Dirichlet Allocation (LDA) - Significance Sentences," *Procedia Comput. Sci.*, vol. 135, pp. 663–670, 2018.
[26] A. Agrawal and U. Gupta, "Extraction based approach for text summarization using k-means clustering," *Int. J. Sci. Res. Publ.*, vol. 4, pp. 1–4, 2014.
[27] M. R. Prathima and H. R. Divakar, "Automatic Extractive Text Summarization Using K-Means Clustering," *Int. J. Comput. Sci. Eng.*, vol. 6, pp. 782–787, Jun. 2018.
[28] H. J. Jain, M. S. Bewoor, and S. H. Patil, "Context Sensitive Text Summarization Using K Means Clustering Algorithm," *Int. J. Soft Comput. Eng.*, vol. 2, pp. 301–304, 2012.
[29] C.-Y. Lin, "ROUGE: A package for automatic evaluation of summaries," *Assoc. Comput. Linguist.*, pp. 74–81, 2004.
[30] K. Papineni, S. Roukos, T. Ward, and W.-J. Zhu, "BLEU: a Method for Automatic Evaluation of Machine Translation," in *Proc. 40th Annu. Meeting Assoc. Comput. Linguist. - ACL '02*, 2001, vol. 371, pp. 311–318.
[31] J. Steinberger and K. Ježek, "Evaluation measures for text summarization," *Comput. Informatics*, vol. 28, pp. 1001–1026, 2009.